\pdfoutput=1

\documentclass[11pt]{article}

\usepackage[hyperref]{acl2021}

\usepackage{times}
\usepackage{latexsym}
\usepackage{arydshln} 
\usepackage{graphicx}
\usepackage{multirow}
\usepackage{subfigure}

\newcounter{example}[section]
\newenvironment{example}[1][]{\refstepcounter{example}\par\medskip
   \noindent \textbf{Example~\theexample. #1} \rmfamily}{\medskip}

\usepackage[T1]{fontenc}

\usepackage[utf8]{inputenc}

\usepackage{microtype}

\aclfinalcopy

\title{Generating Gender Augmented Data for NLP}

\author{{\bf Nishtha Jain\textsuperscript{1}, Maja Popovic\textsuperscript{2},  Declan Groves\textsuperscript{3}, Eva Vanmassenhove\textsuperscript{4}}  \\ \\
  \textsuperscript{1}ADAPT Centre, Trinity College Dublin, \\
  \textsuperscript{2}ADAPT Centre, Dublin City University, \\
  \textsuperscript{3}Microsoft, Dublin, \\
  \textsuperscript{4}Department of CSAI, Tilburg University\\\
  {\small \textls[45] \tt \textsuperscript{1,2}firstname.lastname@adaptcentre.ie, \textsuperscript{3} degroves@microsoft.com,\textsuperscript{4}e.o.j.vanmassenhove@tilburguniversity.edu}\\}

\date{}

\begin{document}
\maketitle
\begin{abstract}

Gender bias is a frequent occurrence in NLP-based applications, especially pronounced in gender-inflected languages. Bias can appear through associations of certain adjectives and animate nouns with the natural gender of referents, but also due to unbalanced grammatical gender frequencies of inflected words. This type of bias becomes more evident in generating conversational utterances where gender is not specified within the sentence, because most current NLP applications still work on a sentence-level context. As a step towards more inclusive NLP, this paper proposes an automatic and generalisable re-writing approach for short conversational sentences. The rewriting method can be applied to sentences that, without extra-sentential context, have multiple equivalent alternatives in terms of gender. The method can be applied both for creating gender balanced outputs as well as for creating gender balanced training data. 
The proposed approach is based on a neural machine translation (NMT) system trained to `translate' from one gender alternative to another. Both the automatic and manual analysis of the approach show promising results for automatic generation of gender alternatives for conversational sentences in Spanish.

\end{abstract}

\section{Introduction}


Recent studies have exposed challenging systematic issues related to bias that extend to a range of AI applications, including Natural Language Processing (NLP) technology \cite{costa2019analysis,blodgett-etal-2020-language}. Observed bias problems range from copying biases already existing in data to claims that the training process can lead to an exacerbation or amplification of
observed biases \cite{Zhou2018, vanmassenhove2021machine}. The algorithms learn to maximize the overall probability of an occurrence, leading to preferences for more frequently appearing training patterns. 


With this work, we propose a method for generating  (more) balanced data in terms of one of the main types of bias frequently observed in language: gender bias.  Gender bias can occur in language due to the fact that some languages have a way of explicitly marking (natural or grammatical) gender while others do not \cite{stahlberg2007representation}. Gender bias in translation is usually manifested when animate entities (e.g. professions) are translated from gender neutral language (e.g. English) into a gendered language (e.g. Spanish) because the instances seen in training data are biased. Also, conversational utterances are prone to bias, both in machine translation as well as in other NLP applications, because systems often do not have the ability to provide multiple gender variants. Therefore, users are simply presented with the most probable option which is prone to bias. In our work, we aim to enable the generation of multiple gender variants by expanding each sentence with the missing gender variants, thus fostering inclusion in online conversations/NLP applications. Generating gender variants can and should also be used to create gender balanced conversational data that can be used to train less biased NLP models such as machine translation models, language models, chat bots, etc.

Unlike previous studies, we did not want to limit ourselves to one specific gender phenomenon, such as   gender markings on professions \cite{zmigrod-etal-2019-counterfactual}) (for which the gender can easily be swapped by using hand-crafted lists) or first person personal pronouns \cite{habash2019automatic}). The objective of this research aims to include as many cases as possible of gender alternatives related not only to gender of persons but also to grammatical gender of the objects referred to. In  
Example~\ref{ex:gender}, (a) illustrates an example of two alternatives for a sentence where there is agreement with the grammatical gender of an object referred to in the previous sentence, while in (b) there is agreement with the gender of the speaker/writer (i.e. a person). 

\begin{example}\\
(a) [MALE] ¿Est\'{a} complet\textbf{o}? -- [FEMALE] ¿Est\'{a} complet\textbf{a}?
\footnote{English: "Is it complete?"}\\
(b) [MALE] Estoy confundid\textbf{o}.  -- [FEMALE] Estoy confundid\textbf{a}.\footnote{English: "I am confused."}
\end{example}\label{ex:gender}

At this stage, our approach does not discriminate between human referents and objects. It is furthermore limited to the generation of binary gender alternatives. We are aware of the importance and challenge of dealing with non-binary gender\cite{ackerman2018syntactic} which we aim to tackle in future work.

The research was carried out in collaboration with an anonymous industry partner with a specific application in mind that deals with conversational sentences. Our approach aims to alleviate gender bias in the said application. We focus on one gender-rich language (Spanish), however, scalability and generalizability were kept in mind while designing the approach.
Our approach can be summarized as follows:


\begin{enumerate}
    \item Identifying (appropriate) sentences/segments that should have the opposite gender variant for some words.  POS sequences were used to extract such segments from the OpenSubtitles corpus\footnote{https://opus.nlpl.eu/}. 
    \item Creating gendered variants for the words in such segments  by applying a rule-based approach. 
    \item Training a neural rewriter on the compiled gender-parallel Spanish data in order to be able to  automatically generate gendered variants on unseen data sets. This additional step makes the approach more scalable as it removes the need for any preprocessing. 
\end{enumerate}

  The first two steps are necessary since there is a lack of readily available open-source gender-parallel data for training. Although language knowledge and a POS tagger are necessary for these steps, the human effort and necessity for external linguistic tools are minimal (contrary to other approaches which heavily rely on linguistic tools~\cite{zmigrod-etal-2019-counterfactual} or on manually created gender-parallel data~\cite{habash2019automatic}. 


\section{Related Work}\label{sec:relatedWork}
In the literature on gender in NLP, two main approaches for bias mitigation can be identified: (a) approaches that attempt to mitigate bias during model or word representation training, and/or (b) approaches that aim to augment the data by creating more variety in the training set (pre-processing step) or in the output (post-processing step). In the following paragraphs, we focus on the latter as it is most closely related to our approach.


There have been attempts to artificially increase the variety in already existing data sets by creating alternatives to sentences in order to decrease the overall bias (in terms of gender).\footnote{Different types of bias exist, however, the current approaches have focused on gender, possibly because many languages have explicit gender markers.} This approach has been referred to in the literature as `Counterfactual Data Augmentation'(CDA)~\citep{Lu2018}. Their CDA approach consists of a simple bidirectional dictionary of gendered words such as he:she, her:him/his, queen:king, etc.  \citet{zhao2018learning} does not use the term CDA as this was introduced later, but what they describe can be interpreted as a rudimentary approach to CDA: they augmented the existing data set by adding additional sentences in which personal pronouns `he' and `she' had been swapped.


Another CDA approach
is described in \citet{zmigrod-etal-2019-counterfactual}. Similar to \citet{Lu2018}, the approach relies on a bidirectional dictionary of animate nouns. Unlike \citet{Lu2018}, pronouns are not handled and the languages worked on are Hebrew and Spanish, languages that have more gender markers than English. Since solely changing the nouns into their male/female counterpart often requires the enforcement of grammatical gender agreement of accompanying articles and adjectives, they introduce Markov Random Fields with optional neural parametrisation that can infer the effect of the swap on the remaining words in the segment. Their approach is limited to mitigating gender stereotypes related to animate nouns and relies on dependency trees, lemmata, POS-tags and morpho-syntactic tags in order to solve issues related to the morpho-syntactic agreement. 

In the field of machine translation (MT), due to specific discrepancies between the information encoded in the source and target data, there has been some work on generating the appropriate gender variant for ambiguous source sentences.\footnote{`I am a teacher' or `I am smart' in English are not marked for gender. However, in many other languages they would be morphologically marked for the male or female gender (e.g. French, Spanish...).} \citet{vanmassenhove2019getting} appends gender tags to the source side of the training data indicating the gender of the speaker. As such, during testing, the desired (or multiple) gender variant(s) can be generated by adding tags. 
\citet{basta2020extensive} also experiment with incorporating a gender tag, and investigate adding the previous sentence as additional context information. Both methods result in the improvement of automatic MT scores as well as on gender accuracy for English-to-Spanish translation. Similarly, \citet{bentivogli2020gender} developed NMT systems using gender tags and evaluated them specifically on gender phenomena.

The work described in \citet{habash2019automatic} is the most similar to ours. They proposed an approach for automatic gender reinflection (``re-gendering") for Arabic. They propose a method which consists of two components: a gender classifier  and a NMT gender rewriter. 
In order to build the NMT rewriter, they first manually created a corpus annotated with gender information. Subsequently, each gendered sentence is re-gendered manually in order to obtain the necessary gender-parallel data for training. This way, they are able to provide gender alternatives for sentences with natural gender agreement with the first person singular. 

Our research, in contrast, aims to augment existing data with gender alternatives in a broader sense: it is not limited to singular first person phenomena, ambiguity in multilingual settings, or phenomena related solely to gender agreement. It involves the gender of adjectives, past participles, and several types of pronouns for which the referent is not explicitly mentioned within the context of the sentence.



\section{Generating gender-parallel data}

As mentioned in the introduction, our main objective is to create an automatic gender rewriter using NMT. In order to do so, we need gender-parallel training data that consists of possible gender variants in both directions (masculine-to-feminine and feminine-to-masculine). Such data sets are, unfortunately, not publicly available, which is why we first leveraged linguistic knowledge and rules to generate a sufficient amount of gender-parallel data. 

Therefore, we identified the sequences of POS classes that show gender agreement in Spanish and can thus be `re-gendered': adjectives, past participles, and several types of pronouns. A detailed description of how the different word classes are tackled to generate gender alternatives is described below. We would like to point out that our target data consisted of very short sentences, where there is at most agreement with one referent.\footnote{For example, sentences such as ``I am happy and they are angry.'' are not covered by our approach as both `happy' and `angry' are in agreement but with different referents, `I' and `they' respectively. Such sentences would require the generation of more than two alternatives since both referents are ambiguous.} As such, our approach is limited to tackle sentences alike and cannot handle the generation of alternatives for sentences where more than two gender alternatives could be generated (due to grammatical agreement of the re-genderable word with multiple entities).

\subsection{Re-genderable word classes}\label{analysedStructures}



\paragraph{Past participles} 
In principle, almost all Spanish past participles have an explicit agreement with their referent and can thus be re-gendered. However, in certain contexts they should not be: if they follow or precede a referent noun (\textit{``Pel\'{i}cula aburrida''}, \textit{``Acceso permitido.''}) thus agreeing with the gender of the noun, or if they  follow the auxiliary verb \textit{``haber''} thus representing past tense and not a property of a person/object (\textit{``he enviado''}, \textit{``has descansado''}). If they appear in isolation (\textit{``Ocupad\textbf{o}/ocupad\textbf{a}.''}, \textit{``Aburrid\textbf{o}/aburrid\textbf{a}.''}), or merely surrounded by interjections or punctuation (\textit{``Ocupad\textbf{o}/ocupad\textbf{a}, gracias.''}, \textit{``Buenos dias, recibid\textbf{o}/recibid\textbf{a}, ¡gracias!''}), adverbs (\textit{``muy cansad\textbf{o}/cansad\textbf{a}''}), or a linking verb (\textit{``Estoy registrad\textbf{o}/registrad\textbf{a}.''}, \textit{``Parece acabad\textbf{o}/acabad\textbf{a}.''}), they can be re-gendered. 

We also included pairs of past participles bound by conjunctions, referring to the same person or object, since in these sentences, both instances should be re-gendered (\textit{``aburrid\textbf{o}/aburrid\textbf{a} y cansad\textbf{o}/cansad\textbf{a}.''}, \textit{``acabad\textbf{o}/acabad\textbf{a} y pagad\textbf{o}/pagad\textbf{a}.''}).

\paragraph{Adjectives}
Many Spanish adjectives are gendered and have an explicit gender marker corresponding to the gender of its referent. However, some adjectives are gender neutral. Gendered and neutral adjectives can (largely) be identified based on their specific suffixes (for example \textit{``-al''}, \textit{``-nte''}, \textit{``-ble''}, so the adjectives \textit{``genial''}, \textit{``interesante''}, and  \textit{``probable''} are neutral), while other suffixes indicate gendered adjectives (for example ``o/a'', so the adjective \textit{``correct\textbf{o}/correct\textbf{a}''} has variants).
 
In addition, similarly to past participles, the given context has to be taken into account for gendered adjectives: they should not be re-gendered if they immediately precede or follow a noun (with or without article) which determines the gender (\textit{``Presupuestos adjuntos.''}, \textit{``¡Maravillosa idea!''}, \textit{``La información correcta.''}). Also, adjectives following neutral demonstrative pronouns \textit{``eso''} or \textit{``esto''} should not be re-gendered (\textit{``Eso es bueno.''}).  Analogous to past participles, adjectives in isolation (\textit{``List\textbf{o}/List\textbf{a}.''}, \textit{``perfect\textbf{o}/perfect\textbf{a}.''}, \textit{``segur\textbf{o}/segur\textbf{a}.''}, \textit{``¡fantástic\textbf{o}/fantástic\textbf{a}!''}), surrounded by punctuation (\textit{``Correct\textbf{o}/correct\textbf{a}, saludos.''}), preceding verb (\textit{``¿Estás list\textbf{o}/list\textbf{a}?''}) or adverb (\textit{``Es muy lind\textbf{o}/lind\textbf{a}.''}) can be re-gendered. 

When two adjectives are present, in a conjunction, and refer to the same referent, both should be re-gendered.


\paragraph{Clitic pronouns} 
Some Spanish clitic pronouns, namely \textit{``lo(s)''} and \textit{``la(s)''} should be re-gendered (e.g. \textit{``L\textbf{o}/l\textbf{a} veo.''}, \textit{``L\textbf{o}/l\textbf{a} adjunto.''}) while \textit{``le(s)''} should not be changed (\textit{``Le veo.''}, \textit{``Le digo.''}). However, in some cases \textit{``lo''} can represent a general concept not referring to a particular object, such as in \textit{``lo siento''} (I’m sorry), \textit{``lo sé''} (I know). If some of these are re-gendered, the precision will decrease.  

\paragraph{Clitic pronouns attached to verbs} 
Clitic pronouns can be attached to a verb infinitive (\textit{``Gracias por acabarl\textbf{o}/acabarl\textbf{a}.''} (thanks for finishing it), \textit{``Quiero verl\textbf{o}/verl\textbf{a}.''} (I want to see it)). Similar to the isolated clitic pronouns, there are certain exceptions, such as \textit{``Es bueno saberlo''} (it is good to know). If the gender neutral clitic pronoun \textit{``le''} is attached to a verb (\textit{``Quiero tenerle informado.''} (I want to keep you/him/her informed)), it should not be re-gendered.
Gendered pronouns attached to an imperative should also be re-gendered (\textit{``Déjal\textbf{o}/Déjal\textbf{a}.''} (leave it), \textit{``Hazl\textbf{o}/Hazl\textbf{a}.''} (do it)). On the other hand, clitic pronouns which refer to an indirect object, such as \textit{``mándame''} (send me), are neutral. Finally, if there are two attached clitic pronouns, \textit{``Mándamel\textbf{o}/Mándamel\textbf{a}.''} (send it to me), only the gendered part (in this case \textit{``lo''/``la''}) should be re-gendered.

\paragraph{Demonstrative pronouns} 
Demonstrative pronouns \textit{``esto''}, \textit{``eso''} and \textit{``aquello''} are neutral, while \textit{``est\textbf{o}s/est\textbf{a}s''}, \textit{``est\textbf{e}/est\textbf{a}''}, \textit{``es\textbf{e}/es\textbf{a}''}, \textit{``aquel\textbf{lo}/aquel\textbf{la}''} are gendered. If the referent is missing in the sentence and the pronoun is gendered, they should be re-gendered. 

\subsection{Adding gender variants by rules}

Whether a gender alternative translation should be generated does not solely depend on the word classes it contains but also on the structure of the sentence. If the referent is missing in a sentence, then an additional variant with the opposite gender should be generated. If the referent is present in a sentence, only one gender variant is grammatically correct, and as such, these sentences are to be left unchanged. The presence or absence of a referent can be determined by the sequence of POS tags in a sentence\footnote{Assuming that the sentences are short- this approach would not generalize to longer sentences}. For example, if we want to check whether a sentence with an adjective ``creo que es correct\textbf{a}" (gloss: ``I believe (it) is correct-feminine") needs an additional re-gendered variant or not, its POS sequence ``VERB CONJUNCTION VERB ADJECTIVE" indicates that there is no referent noun within the given context. Therefore, another variant of the adjective ``correct" should be provided: ``creo que es correct\textbf{o}". In contrast, the sentence ``la soluci\'{o}n es correcta" with POS sequence ``ARTICLE NOUN VERB ADJECTIVE" contains a referent noun ``soluci\'{o}n", and therefore it should not be re-gendered.
 

For each re-genderable sentence, we apply rules for changing the ending of the corresponding word, if necessary.
The POS sequences to identify re-genderable sentences and the subsequent rules used to re-gender the corresponding words in such sentences are given in detail in the Appendix. It is worth mentioning  we also used POS sequences to identify neutral sentences (those which should be not re-gendered ) since we wanted the parallel corpus to contain both.

\section{Gender-parallel data}
In order to create gender-parallel data, a set of Spanish subtitles was downloaded from the OPUS~\cite{Tiedemann2012}  website.\footnote{http://opus.nlpl.eu/} After basic filtering (removing too long and non-alpha numeric segments), a set of short sentences with up to 10 (untokenized) words was extracted. This candidate set consisted of 22 458 968 sentences. This data set was POS tagged using Treetagger\footnote{https://www.cis.uni-muenchen.de/~schmid/tools/TreeTagger/}. The sentences matching the POS sequences mentioned in the Appendix were extracted from this data set. This set consisted of more than 1M sentences. For each extracted re-genderable sentence, the alternative gender variant is created by applying appropriate rules described in the Appendix. After applying rules on all re-genderable structures, we joined both re-gendering directions (masculine-to-feminine and feminine-to-masculine) in order to create a balanced data set. As already mentioned, the corpus also contains a number of sentences that are not to be regendered. 
 By including these neutral sentences in our training data, we encourage the rewriter to: (a) learn when to generate alternatives and when not to, and (b) how to generate those alternatives, if necessary. 
In this way, a corpus with about 2.2M gender-parallel sentences was created. This corpus was then separated into train, development ($\sim$1k sentences) and test ($\sim$3k sentences) sets. The rewritten parts of the development and test sets were revised manually and the errors were corrected for about 6\% of sentences and 1.5\% of words. The training set, being large, was not verified manually, thus it contained some noise.

In addition to OpenSubtitles, we also obtained data from the industry partner consisting of around 8 000 sentences readily available with all possible alternative versions of the sentences provided. An additional 22 000 sentences had to be revised manually in order to produce the correct gender variant for re-genderable sentences. This set was used as an additional test set for the re-writer. One part of this set can be handled by the described POS sequences and rules (``structured test 1"), while another part contains different POS sequences and cannot be handled by these rules at all (``unstructured test 1"). The latter test set will give a good estimation of the scalability of our approach. An overall split of data sets is described in  Table~\ref{tbl:data-stat}. The OpenSubtitles data was split in the standard way for machine translation, namely a few thousands of segments for development and test sets and the rest for the training set.

\begin{table}[ht]
    \small
    \centering
    {\setlength\tabcolsep{4pt}\begin{tabular}{|l||r|}\hline
    set & segments \\ \hline
    \hline
       training (OpenSubtitles)  &  2 193 657 \\
                 \hline
        development (OpenSubtitles) & 1 018 \\
                    \hdashline

        test (OpenSubtitles) & 3 066 \\
             \hline
    structured test1  & 5 648 \\
          \hline
    unstructured test1 & 15 892 \\
          \hline
    \end{tabular}}
    \caption{Statistics of data used for building the NMT rewriter.}\label{tbl:data-stat}
    \label{tbl:stat}
\end{table}

\section{Neural Rewriter}\label{sec:nmtRewriter}
Once we compiled a sufficient amount of gender-paralell data, we were able to train our automatic rewriter. 
The automatic rewriter is a NMT system trained on the following parallel data: original sentences as the source language, and re-gendered sentence as the target language. For neutral sentences, the source and the target parts are identical.  

The NMT rewriter was built using the publicly available Sockeye\footnote{\url{https://github.com/awslabs/sockeye}} implementation~\cite{hieber-etal-2018-sockeye} of the Transformer architecture~\cite{Vaswani2017}. The system operates on sub-word units
generated by byte-pair encoding (BPE)\cite{sennrich-etal-2016-neural}. We set the number of BPE merging operations to 32000. We have experimented with the following setups:
\begin{itemize}
\item a Standard NMT system without any additional tags
\item an NMT system with neutrality/re-genderability tags in the source part 
\end{itemize}
The system with tags was built using the same technique as proposed in~ \cite{GoogleZST} for multilingual MT systems and used for many other applications including gender-informed MT~\cite{vanmassenhove2019getting}. For our experiments, we added a label `N' (neutral) or `G' (re-genderable) to each source sentence.
These tags are implicitly present in the gender-parallel data -- if the source and the target parts differ, it is a re-genderable sentence, if they are identical it is neutral. Therefore, the tags are certainly available for the training and development sets, but they might not be available for the test sets.
Therefore, this system was assessed in two ways:
\begin{itemize}
\item ``NMT-T'': neutrality/re-genderability tags are available for the test sets 
\item ``NMT-AT'': the tags are not available for the test sets (a realistic scenario) and therefore are assigned automatically by the gender classifier described in the next section (which is similar to the approach described in~\cite{habash2019automatic}.)
\end{itemize}

\subsection{Gender Classifier}\label{sec:class} 

In order to explore potential benefits of automatic  pre-classification for automatic rewriting,
a classifier to distinguish between `re-genderable' (G)\footnote{Grammatical gender markings are not related to a referent within the sentence, therefore these markings have to be expanded.} and `neutral' (N)\footnote{No gender markers that need to be expanded.} sentences was also designed. The tags generated by this classifier were used to assess the performance of the ``NMT-AT'' re-writer by appending them to the sentences. 

\subsubsection*{Data}\label{classifierData}
The classifier was built on the data set of about 8 000 sentences provided by the industry partner. These sentences were balanced in both directions i.e., both masculine-to-feminine as well as feminine-to-masculine counterparts of a given sentence were present and labelled as G. The rest of the sentences were labeled as N.

For the sake of designing a generalised classifier, the development set consisted of sentences from the OpenSubtitles corpus (and was the same as the development set used for the NMT system).

The final classifier was tested on two different test sets - one consisted of the 22 000 conversational sentences sourced from the industry partner and another extracted from the OpenSubtitles corpus.

\subsubsection*{Features}

Following on the work of \citet{habash2019automatic} for the gender identification step, features using character $n$-grams, word $n$-grams and morphological information were created from the training data. To begin with, TF-IDF scores of character $n$-grams of length 4-7 with maximum features capped at 20 000 and of word $n$-grams of length 1-3 were generated. These two feature matrices were joined together along with a morphological feature that denoted the presence of a gendered word in the sentence. The resulting training data was a high dimensional data frame with around 40 000 features.

Due to the limited size of the training set, neural network based classifiers were ruled out. Instead, owing to the high dimensional nature of the data, we used a SVM based classifier for training. All the steps described in this section were implemented in Python 3.7 using sklearn\footnote{https://scikit-learn.org/stable/}, pandas\footnote{https://pandas.pydata.org/} and StanzaNLP\footnote{https://stanfordnlp.github.io/stanza/} libraries.

\subsubsection*{Precision and Recall}

The SVM based classifier was tested on two sets of data as described in Section \ref{classifierData}. This was done in order to assess the generalisability of the classifier. Given the small size of the training data, the performance of the classifier looks promising thus far (see Table \ref{tbl:classifierResults}). 

It can be observed in Table \ref{tbl:classifierResults} that the classifier clearly performs better on the test data set consisting of sentences sourced from the industry partner as compared to the data extracted from OpenSubtitles. While the accuracy is comparable on both sets (~80\%), the precision and recall of neutral sentences is higher on the industry data than the set compiled from OpenSubtitles data. The high recall of sentences labelled as G implies that the classifier is almost always successful at recognising sentences that need to be re-gendered (i.e. sentences that need an alternative variant). However, it incorrectly predicts the labels of a substantial number of N-labelled sentences, which in turn results in a low precision of re-genderable sentences. As we want to avoid generating (incorrect) gender alternatives for neutral sentences, our aim was to first attain a high precision for neutral sentences and then aim towards a high recall for the same. The tags generated by this classifier for the industry sourced data and OpenSubtitles data were used to test the ``NMT-AT'' rewriter.

\begin{table}
    \centering
    {\setlength\tabcolsep{2.5pt}\begin{tabular}{|c||c|c|c|c|c|c|}\hline
    {} & \multicolumn{3}{c|}{Industry Test Set} & \multicolumn{3}{c|}{OpenSubs} \\ \cline{2-7}
    {} & Acc. & Rec. & Prec. & Acc. & Rec. & Prec. \\\hline \hline
    Overall & 82\% & - & - & 80\% & - & -\\
    G & - & 96\% & 60\% & - & 97\% & 76\% \\
    N & - & 76\% & 98\% & - & 56\% & 93\% \\\hline
    \end{tabular}}
    \caption{Gender Classifier Results}\label{tbl:classifierResults}
\end{table}

\section{Results for generating gender variants}

Our first experiment consisted of using the implementation of CDA by \cite{zmigrod-etal-2019-counterfactual} to generate gendered variants. However, this work only tackled animate nouns, which rarely occur in the conversational sentences we investigated in this work. Our re-implementation of their approach generated the correct gender variant for only ~1\% of the sentences. Because of the very low recall, this implementation was not directly applicable for our research. In addition to this, since our work aims to tackle multiple gender related word classes, we explored extending the implementation by augmenting the list with character adjectives. On doing so, we found that this implementation generated the correct gendered variant in only ~9\% of the cases. An important point to note is that 3\% of the neutral sentences (for which variants should not have been generated) were also converted as opposed to the 1\% with only animate nouns, attributed to the presence of more words in the hand-crafted lists. In order to cover more words and improve the performance of this implementation on our data set, we considered augmenting the hand-crafted list with past participles and/or clitic pronouns. However, that increased the size of the list exponentially and made the approach prone to errors, inefficient and not scalable to other languages.

\subsection{Automatic evaluation of neural rewriter} 

The results in the form of error rates are shown in~\ref{tbl:results}. Since we are not performing typical machine translation, namely converting one language into another one, but only converting a few words in the sentence into a sentence in the same language, these error rates are not related to any of the typical automatic evaluation metrics (such as TER, etc.) but to the amount of incorrectly converted words.
For each system, numbers in the left column represent the count of incorrectly converted words normalised by the total number of sentences, while numbers in the right column represent the count of incorrectly converted words normalised by the total number of words in the corpus. 
The numbers in the first row and first two columns can be interpreted as follows: left: 6.4\% of all sentences have incorrectly converted words in ; right: 1.50\% of all words are incorrectly converted.

\begin{table*}[tbh!]
    \centering
    \begin{tabular}{|l|l||cc|cc|cc|cc|}\hline
    set & type & \multicolumn{2}{c|}{NMT} & \multicolumn{2}{c|}{NMT-T} & \multicolumn{2}{c|}{NMT-AT} & \multicolumn{2}{c|}{rules} \\
    \hline \hline
    test     & all &  6.4 & 1.50 & 4.5 & 1.03 & 17.9 & 4.21 & 6.1 & 1.43 \\
    (structured)         & neutral & 5.3 & 1.13 & 2.5 & 0.48 & 33.3 & 7.07 & 0.0 & 0.0  \\
             & re-genderable & 7.1 & 1.81 & 6.0 & 1.51 & 6.0 & 1.72 & 6.1 & 1.43 \\ 
             \hline
test1 & all & 2.4 & 0.54 & 1.3 & 0.27 & 4.5 & 0.99 & 3.2 & 0.7 \\
(structured)      & neutral & 4.8 & 0.95 & 2.2 & 0.43 & 8.7 & 1.73  & 0.0 & 0.0 \\
      & re-genderable & 0.8 & 0.19 & 0.6 & 0.14 & 1.6 & 0.38 & 3.2 & 0.7 \\ 
             \hline \hline
 test2 & all & 11.9 & 2.13 & 5.2 &  0.93 & 10.4 & 1.87 & \multicolumn{2}{c|}{not} \\
(unstructured) & neutral & 3.3 & 0.58 & 0.3 & 0.04  & 6.0 & 1.07 & \multicolumn{2}{c|}{applicable} \\
         & re-genderable & 57.3 & 10.7 & 31.1 & 5.84  & 33.4 & 6.26 & \multicolumn{2}{c|}{} \\
             \hline
    \end{tabular}
    \caption{Results for NMT rewriter: error rates (\%): count of incorrectly converted words normalised by the total number of sentences (left columns) and normalised by the total number of words (right columns).  }
    \label{tbl:results}
\end{table*}

First, it can be noted that the error rates are lower for the template-based ``in-domain'' test sets than for the unstructured ``out-of-domain'' test sets, which is in line with our expectations. The change in error rate is mainly due to discrepancies in the re-genderable segments. The error rates in the neutral segments are comparable in the out-of-domain and in-domain test sets.

Adding manual tags indicating whether a sentence should get a gender alternative or not (e.g. ‘neutral’ vs ‘regenderable’) reduces the error rates on all test sets for both types of segments. A similar performance can not be achieved by adding automatic tags. Automatic tags deteriorate the performance on neutral segments, but reduce the error rates for re-genderable segments, especially for the unstructured ``out-of-domain'' test set. The manually tagged results indicate the potential of a classifier. These results tie up with the results of the gender classifier (Section~\ref{sec:class}) which is good at classifying the re-genderable sentences as denoted by a high recall of sentences labelled ‘G’, however it doesn't do very well at labelling neutral sentences as ‘N’. It tends to mislabel many of those sentences as ‘G’, resulting in a low recall and, consequently, incorrect re-gendering.

For the sake of completeness, error rates are reported for the rule-based rewriter, too. The error rates for re-genderable sentences are lower than the NMT rewriter without tags and for neutral sentences the error rate is 0\%; it should be noted that the rules are applicable only to data sets which strictly conform to the described template structures. 

\subsection{Qualitative manual inspection of errors}
In order to better understand the nature of errors and remaining challenges, a qualitative manual inspection was carried out on all test sets. First of all, it is observed that in general, the NMT re-writer does not intervene on large portions of a sentence but addresses only specific words, which is exactly what it is expected to do. This is a positive result, as generating gender variants implies changing specific gendered words and does not involve changing entire segments. It also facilitates the evaluation since manual inspection is needed only to identify the nature of incorrect words. 

\begin{table*}
\centering
\subtable[structured sentences]{
\begin{small} 
\begin{tabular}{|l|l|l|l|l|}\hline
type & original     & correct & NMT & NMT-T \\
\hline
  N     & esto es perfecto &	esto es perfecto 	& esto es {\bf perfecta}	& esto es perfecto \\
  \hline
 G & est\'{a} adjunt\textbf{o} & 	est\'{a} adjunt\textbf{a}  & 	est\'{a} adjunt\textbf{o} & 	est\'{a} adjunt\textbf{o}  \\
 \hline
\end{tabular}
\end{small} 
}

\vspace{1cm}
\subtable[unstructured sentences]{ 
\begin{small} 
\begin{tabular}{r|l|l|l|l|l|}\cline{2-6}
 & type & original     & correct & NMT & NMT-T \\
\hline
1) & N &	no son lo mismo & no son lo mismo & 	no son {\bf la misma} &	no son lo mismo \\
 \hdashline
2) & N & 	aquello fue encantador & 	aquello fue encantador &	aquello fue {\bf encantadora} &	aquello fue encantador \\
 \hdashline
3) & N &	¿a qui\'{e}n aprovecha?	& ¿a qui\'{e}n aprovecha?	& ¿a qui\'{e}n {\bf aprovecho}?	 &¿a qui\'{e}n aprovecha? \\
\hdashline
4) & N &	ind\'{i}queme la  & ind\'{i}queme la  &	ind\'{i}queme la  & 	ind\'{i}queme la  \\
& &	disponibilidad &  disponibilidad &	{\bf emperbilidad} &  {\bf evelbilidad}  \\
\hdashline

 5) & N	& ind\'{i}queme su   & 	ind\'{i}queme su  & 	ind\'{i}queme su  &	ind\'{i}queme su  \\
 & & disponibilidad & 	disponibilidad & 	 disponibilidad &	{\bf escorpibilidad}  \\
 \hdashline
6) & N & unos momentos & 	unos momentos  &	unos momentos  &	unos momentos \\
& & extraordinarios & 	 extraordinarios &	 {\bf extraordinarias arios} &	extraordinarios \\
\hdashline
7) & N &	ind\'{i}quenos cu\'{a}nto &	ind\'{i}quenos cu\'{a}nto &	{\bf ind\'{i}quenas} cu\'{a}nto & 	ind\'{i}quenos cu\'{a}nto \\
\hline
 8) & G & esta es la adecuada & este es el adecuado & {\bf esta} es {\bf \em la} {\bf adecuada} & {\bf esta} es {\bf \em lo} {\bf adecuada}  \\
 \hdashline
9) & G & esta la hemos recibido  &         este lo hemos recibido &        {\bf esta la}  hemos recibido &       {\bf esta}  lo hemos recibido \\
\hline
\end{tabular}
\end{small} 
    }
\caption{Examples of incorrectly generated sentence variants for (a) structured sentences  and (b) unstructured sentences. }\label{tbl:examples}
\end{table*}


The analysis revealed that the most frequent error for neutral sentences are re-gendered pronouns and adjectives which should not be changed. 
Also, the most frequent error in re-genderable sentences is leaving them unchanged. 
These types of errors are predominant in structured sentences, and two examples, one for neutral and one for regenderable sentence, can be seen in Table~\ref{tbl:examples}(a). It can also be seen that adding tags can help in some cases. 

For unstructured sentences, there are more error types especially for neutral sentences, and examples can be seen in Table~\ref{tbl:examples}(b). 
In the first three sentences, the same error type as for structured sentences can be seen, namely some words are changed which should not be changed. Adding tags helped in both cases. 
However, some other error types can be seen, such as converting some (not gender-related) words into non-existing words in sentences 4) and 5). For sentence 5), generating a non-existing word was triggered by adding tags. 
Sentence 6) shows an unnecessary re-gendering as well as adding non-existing words. This was also resolved by adding tags.
In sentence 7), a word which is not at all related to gender was converted, and this was prevented by adding tags.

As for regenderable sentences, the vast majority of errors are again the unchanged words which had to be changed. If there is more than one word to be regendered, sometimes they all remain unchanged (sentence 8) and sometimes only some of them are regendered (sentence 9). Tags can help to some extent, but only for some words, not all. 


Adding tags generated by the classifier also increases the number of correctly re-gendered structures at the cost of a small number of additions of non-existing words.

\section{Conclusions and Future Work} \label{conclusion}

In this paper, we describe an initial approach towards enriching short conversational sentences with their gender variants.  Unlike other related work, our approach is not limited to tackling the first person singular phenomena, swapping third person pronouns or merely dealing with occupational or generally animate nouns. In addition, with our approach, the reliance on linguistic knowledge and tools is kept to a minimum in order to facilitate real-world deployment.

The main hurdle for this type of research is the absence of large training sets. Although provided with some manually annotated data from the industry partner, the data provided was far from sufficient to train a state-of-the-art automatic gender re-writer. 

Therefore, training data was extracted from OpenSubtitles using linguistic knowledge about the targeted language, namely Spanish. Re-genderable types of words (POS classes) were identified and then frequently occurring `re-genderable' as well as 'neutral' POS patterns were extracted. By applying the corresponding rules to the re-genderable sentences, a large gender-parallel Spanish data set was compiled.

Next, an NMT rewriter was trained in order to `translate' each re-genderable sentence into its gender alternative which showed promising performance both in terms of automatic as well as of manual evaluation. 

In addition, it is shown that providing additional information regarding the need for rewriting in the form of tags could be helpful for the NMT system, as similar tags have shown to be useful for other applications such as multilingual translation, controlling politeness and gender in MT, etc.  While gold standard labels show better performance than the labels generated by the gender classifier, the classifier shows promising results given the very small training set.  Further experiments should investigate a classifier trained on larger amount of data. 
 
In future work, we would like to explore how a similar approach can be applied on more sentence structures in Spanish, as well as for different languages which exhibit distinct gendering rules. Furthermore, different NMT architectures, e.g. character-level NMT or an NMT system with linguistically motivated subword units  could be an interesting extension to the conducted experiments, given that gender is usually marked by specific morphemes (usually not more than one or two specific characters). In addition to that, the performance of the gender classifier can be improved to produce more accurate tags by using larger annotated training sets, adding more morphological information in features and using word embeddings instead of TF-IDF scores.





\bibliography{naacl2021}
\bibliographystyle{acl_natbib}

\end{document}


\maketitle

\section{Appendix}\label{sec:appendix}

\subsection{Training Data Split}

This section provides a detailed split of the data used in this research work, including OpenSubs corpus and the data obtained from our industry partner. Table \ref{tbl:data-stat-details} mentions the split of neutral and re-genderable segments in each set.

\begin{table}[htbp]
    \small
    \centering
    {\setlength\tabcolsep{4pt}\begin{tabular}{|l|l||r|r|}\hline
     &  & \# of  & \# of running \\
    set & type & segments & words \\ \hline
    \hline
       training  &  all & 2 193 657 & 15 540 108 \\
       for NMT          & neutral & 1 145 781 & 8 360 144 \\
    rewriter             & re-genderable & 1 047 876 & 7 179 964 \\
                 \hline
        development & all & 1 018 & 4 350 \\
                  & neutral & 432 & 2 021 \\ 
                    & re-genderable & 506 & 2 329 \\ 
                    \hdashline

        test & all & 3 066 & 12 996 \\
             & neutral & 1 289 & 6 045 \\ 
             & re-genderable & 1 777 & 6 951 \\ 
             \hline
    structured  & all & 5 648 & 25 605 \\
    test1 & neutral & 2 304 & 11 607 \\
          & re-genderable & 3 344 & 13 998\\
          \hline
    unstructured & all & 15 892 & 88 537  \\
    test1 & neutral & 14 497 & 81 075 \\
          & re-genderable & 2 752 & 14 680 \\ 
          \hline
training & all & 7 692 & 36 250 \\
for      & neutral & 3 440  & 17 438\\ 
 classifier     & re-genderable & 4 252 & 18 812\\ 
    \hline
    \end{tabular}}
    \caption{Data statistics}\label{tbl:data-stat-details}
    \label{tbl:stat-det}
\end{table}

\subsection{POS Sequences and Rewriting Rules}\label{rules}

This section mentions in detail the word categories analyzed in the industry sourced data. The corresponding POS sequences and rules formulated to rewrite gender variants for those particular word categories are given in respective tables.

\subsubsection*{Clitic Pronoun Candidates}
Table \ref{tab:clitic_pron_g} consists of the POS sequences of gendered utterances that contain past clitic pronouns.

\begin{table}[htbp]
    \small
    \centering
    {\setlength\tabcolsep{4pt}\begin{tabular}{|l||l|} \hline
    POS Sequences including  &    Rewriting Rules  \\ regenderable clitic pronouns (PPC)   &   for each PPC    \\ \hline \hline
    PPC-Vfin-FS   &  \\ \cline{1-1}
    FS-PPC-Vfin-FS &  \\ \cline{1-1}
    PPC-Vfin-ADV-FS  & \\ \cline{1-1}
    Vfin-CQUE-PPC-Vfin-FS  &  \\ \cline{1-1}
    NEG-PPC-Vfin-FS  &  \\  \cline{1-1}
    ADV-PPC-Vfin-FS  & ``lo'' => ``la''  \\  \cline{1-1}
    ADV-CM-PPC-Vfin-FS  & ``la'' => ``lo''  \\  \cline{1-1}
    PPC-Vfin-CM-NC-FS   &  ``los'' => ``las''  \\  \cline{1-1}
    ADV-NEG-PPC-Vfin-FS  &  ``las'' => ``los'' \\  \cline{1-1}
    NEG-PPC-Vfin-ADV-FS  &   \\  \cline{1-1}
    Vfin-CQUE-NEG-PPC-Vfin-FS  &   \\  \cline{1-1}
    NEG-Vfin-CQUE-PPC-Vfin-FS  &   \\  \hline
    \end{tabular}}
    \caption{POS sequences and rewriting rules for clitic pronouns}
    \label{tab:clitic_pron_g}
\end{table}

\subsubsection*{Demonstrative Pronoun Candidates}
Table \ref{tab:dem_pronoun_g} consists of the POS sequences of gendered utterances that contain demonstrative pronouns.

\begin{table}[htbp]
    \small
    \centering
    {\setlength\tabcolsep{4pt}\begin{tabular}{|l||l|} \hline
    POS Sequences  including  &  Rewriting Rules  \\
    regenderable  demonstrative &   for each DM    \\ 
     pronouns (DM)&      \\ 
    \hline \hline
    Vfin-DM-FS   & ``este'' => ``esta'' \\ \cline{1-1}
    FS-Vfin-DM-FS & ``esta'' => este'' \\ \cline{1-1}
    FS-INT-Vfin-DM-FS  & ``estas'' => ``estos'' \\ \cline{1-1}
    NEG-Vfin-DM-FS  &  ``ese'', => ``esa'' \\ \cline{1-1}
    FS-NEG-Vfin-DM-FS  &  ``esa'' => ``ese'' \\  \cline{1-1}
    DM-FS  & ``esos'' => ``esas'' \\  \cline{1-1}
    DM-SE-Vfin-FS  & ``esas'' => ``esos''  \\  \cline{1-1}
    ADV-Vfin-DM-FS   &  ``aquel'', => ``aquella''  \\  \cline{1-1}
    DM-NEG-FS  &  ``aquella'' => ``aquel'' \\  \cline{1-1}
    DM-PPX-Vfin-FS  &  ``aquellos'' => ``aquellas'' \\  \cline{1-1}
    FS-CC-DM-FS  &  ``aquellas'' => ``aquellos'' \\  \cline{1-1}
    DM-NEG-Vfin-FS  &  ``estos'' => ``estas'' \\  \hline
    \end{tabular}}
    \caption{POS sequences and rewriting rules for demonstrative pronouns}
    \label{tab:dem_pronoun_g}
\end{table}

\subsubsection*{Past Participles Candidates}
Table \ref{tab:past_participle_g} consists of the POS sequences of gendered utterances that contain past participles.

\begin{table}[htbp]
    \small
    \centering
    {\setlength\tabcolsep{4pt}\begin{tabular}{|l||l|} \hline

    POS Sequences  including &  Rewriting Rules   \\
    regenderable past participles (Vadj) & for each Vadj      \\ \hline \hline
    Vadj-FS   & if word suffix is \\ \cline{1-1}
    Vfin-Vadj-FS & ``ado'', ``ido'', ``cho''  \\ \cline{1-1}
    Vadj-CC-Vadj-FS  & => last letter to ``a'' \\ \cline{1-1}
    Vfin-ADV-Vadj-FS  & if the word suffix is \\ \cline{1-1}
    FS-Vfin-Vadj-FS  & ``ada'', ``ida'', ``cha''  \\  \cline{1-1}
    FS-Vadj-FS  & => last letter to ``o''  \\  \cline{1-1}
    ADV-Vadj-FS  & if word suffix is  \\  \cline{1-1}
    ADV-Vfin-DM-FS   &  ``ados'', ``idos'', ``chos''  \\  \cline{1-1}
    Vadj-ADV-FS  & => last two letters ``as'' \\  \cline{1-1}
    FS-Vfin-ADV-Vadj-FS  & if the word suffix is \\  \cline{1-1}
    ADV-CM-Vadj-FS  & ``adas'', ``idas'', ``chas''\\ \cline{1-1} 
    ADV-Vfin-Vadj-FS  & =>  last two letters ``os'' \\  \cline{1-1}
    NEG-Vadj-FS  &  \\  \hline
    \end{tabular}}
    \caption{POS sequences and rewriting rules for past participles}
    \label{tab:past_participle_g}
\end{table}

\subsubsection*{Adjective Candidates}
Table \ref{tab:adjective_g} consists of the POS sequences of gendered utterances that contain adjectives.

\begin{table}[htbp]
    \small
    \centering
    {\setlength\tabcolsep{4pt}\begin{tabular}{|l||l|} \hline

    POS Sequences including & Rewriting Rules \\
    regenderable adjectives (ADJ) & for each ADJ \\ \hline \hline
    ADJ-FS & \\ \cline{1-1}
    Vfin-ADJ-FS & \\ \cline{1-1}
    FS-Vfin-ADJ-FS &  \\ \cline{1-1}
    ADV-ADJ-FS & if suffix ``o'' => ``a'' \\ \cline{1-1}
    Vfin-ADV-ADJ-FS & if suffix ``dor'' => ``dora'' \\  \cline{1-1}
    FS-ADJ-FS & if suffix ``os'' or ``dores'' =>  \\  \cline{1-1}
    FS-Vfin-ADV-ADJ-FS & last two letters to ``as'' \\ \cline{1-1}
    ADV-Vfin-ADJ-FS & if suffix ``dora'' => ``dor'' \\  \cline{1-1}
    NEG-Vfin-ADJ-FS & if suffix ``doras'' => ``dores'' \\  \cline{1-1}
    FS-INT-ADJ-FS & if suffix ``a'' => ``o'' \\  \cline{1-1}
    VMfin-Vinf-ADJ-FS & if suffix ``as'' => ``os'' \\  \cline{1-1}
    SE-Vfin-ADJ-FS & \\  \cline{1-1}
    ADJ-CC-ADJ-FS & \\  \hline
    \end{tabular}}
    \caption{POS sequences and rewriting rules for adjectives}
    \label{tab:adjective_g}
\end{table}

\subsubsection*{Clitic Pronouns Attached to Verbs}
Table \ref{tab:vcl_rules} consists of the rewriting rules applied to gendered utterances that contain clitic pronouns attached to verbs.

For clitic pronouns attached to verbs, if a VCL tag is present in the POS sequence of the sentence then it represents a VCL candidate\footnote{POS tags for this category are not very clean, many of verbs with clitic pronouns are tagged as a simple verb infinitive, therefore this rule was included (infinitives without clitic pronouns cannot end with ``lo/la/los/las'').}. Table \ref{tab:vcl_rules} represents the rules to tackle such structures.

\begin{table}[htbp]
    \small
    \centering
    {\setlength\tabcolsep{4pt}\begin{tabular}{|l|} \hline
    Rewriting Rules for each clitic \\ 
    pronoun attached to a verb \\
   \hline \hline
    if suffix ``lo'' => ``la'' \\ \hline \
    if suffix ``la'' => ``lo'' \\ \hline \
    if suffix ``los'' => ``las'' \\ \hline \
    if suffix ``las'' => ``los'' \\ \hline
    \end{tabular}}
    \caption{POS sequences and rewriting rules for clitic pronouns attached to verbs}
    \label{tab:vcl_rules}
\end{table}

\subsubsection*{Neutral Past Participle Structures}
Table \ref{tab:past_participle_n} consists of the POS sequences which contain past participles which should not be regendered.

\begin{table}[htbp]
    \small
    \centering
    {\setlength\tabcolsep{4pt}\begin{tabular}{|l|} \hline
    POS Sequences including past participles \\ (Vadj) which should not be regendered \\ \hline \hline
    NC-Vadj-FS \\ \cline{1-1}
    FS-NC-Vadj-FS \\ \cline{1-1}
    Vadj-CC-Vadj-FS \\ \cline{1-1}
    VHfin-Vadj-ART-NC-FS \\ \cline{1-1}
    FS-NC-Vadj-FS \\  \cline{1-1}
    ART-NC-SE-VHfin-Vadj-FS  \\  \cline{1-1}
    ART-NC-Vfin-Vadj-FS \\ \cline{1-1}
    ADV-Vadj-NC-FS \\  \cline{1-1}
    FS-ADV-Vadj-NC-FS \\  \cline{1-1}
    Vfin-ADV-Vadj-NC-FS \\  \cline{1-1}
    FS-Vfin-ADV-Vadj-NC-FS \\  \hline

    \end{tabular}}
    \caption{POS sequences containing past participles which should not be regendered}
    \label{tab:past_participle_n}
\end{table}

\subsubsection*{Neutral Adjective Structures}
Table \ref{tab:adjective_n} consists of the POS sequences containing adjectives which should not be regendered.

\begin{table}[htbp]
    \small
    \centering
    {\setlength\tabcolsep{4pt}\begin{tabular}{|l|} \hline
    POS Sequences including adjectives \\ 
    (ADJ) which should not be regendered \\ \hline \hline
    FS-ADJ-NC-FS \\ \hline
    Vfin-ART-NC-ADJ-FS \\ \hline
    FS-Vfin-ART-ADJ-NC-FS \\ \hline
    FS-INT-ADJ-NC-FS \\ \hline
    NC-ADJ-FS \\  \hline
    ART-NC-Vfin-ADJ-FS  \\  \hline
    ADV-ADJ-NC-FS \\ \hline
    FS-ADV-ADJ-NC-FS \\  \hline
    Vfin-ADV-ADJ-NC-FS \\  \hline
    FS-Vfin-ADV-ADJ-NC-FS \\  \hline
    \end{tabular}}
    \caption{POS sequences containing adjectives which should not be regendered}
    \label{tab:adjective_n}
\end{table}